\documentclass{article}

\usepackage[dblblindworkshop, final]{neurips_2025}
\workshoptitle{Deep Learning for Code in the Agentic Era}

\usepackage[utf8]{inputenc} % allow utf-8 input
\usepackage[T1]{fontenc}    % use 8-bit T1 fonts
\usepackage{hyperref}       % hyperlinks
\usepackage{url}            % simple URL typesetting
\usepackage{booktabs}       % professional-quality tables
\usepackage{amsfonts}       % blackboard math symbols
\usepackage{nicefrac}       % compact symbols for 1/2, etc.
\usepackage{microtype}      % microtypography
\usepackage{xcolor}         % colors

\usepackage{amsmath}
\usepackage{booktabs}
\usepackage{graphicx}
\usepackage{float}
\usepackage{xcolor}
\usepackage{soul}
\usepackage{enumitem}
\usepackage{wrapfig}

\title{Practical Code RAG at Scale: Task-Aware Retrieval Design Choices under Compute Budgets}
\author{
Timur Galimzyanov$^1$, Olga Kolomyttseva$^{1}$\thanks{Work done while internship at JetBrains Research.}\kern0.2em, Egor Bogomolov$^{1,2}$ \\
$^1$JetBrains Research, $^2$Delft University of Technology\
\texttt{timur.galimzyanov@jetbrains.com}
}

\date{}

\begin{document}

\maketitle

\begin{abstract}
% Old version
% We evaluate retrieval-augmented generation (RAG) configurations for code completion and bug localization tasks. We investigate how different components — chunking strategies, scoring methods, and text-splitting approaches — affect performance across various context window sizes. For code completion, BM25 with word-level splitting outperforms other techniques, while bug localization benefits significantly from dense embeddings. We demonstrate that optimal chunk size correlates with model context capacity, with smaller models ($\geq$4000 tokens) performing best with 32-64 line chunks. Surprisingly, simple line-based chunking performs competitively with more complex language-aware approaches. Our latency analysis reveals efficiency differences of up to 200× between methods. Thus, we provide evidence-based recommendations for implementing effective code-oriented RAG systems based on task requirements, model constraints, and computational efficiency. 

We study retrieval design for code-focused generation tasks under realistic compute budgets. Using two complementary tasks from Long Code Arena — code completion and bug localization — we systematically compare retrieval configurations across various context window sizes along three axes: (i) chunking strategy, (ii) similarity scoring, and (iii) splitting granularity. (1) For PL→PL, sparse BM25 with word-level splitting is the most effective and practical, significantly outperforming dense alternatives while being an order of magnitude faster. (2) For NL→PL, proprietary dense encoders (Voyager-3 family) consistently beat sparse retrievers, however requiring ~100x larger latency. (3) Optimal chunk size scales with available context: 32–64 line chunks work best at small budgets, and whole-file retrieval becomes competitive at 16000 tokens. (4) Simple line-based chunking matches syntax-aware splitting across budgets. (5) Retrieval latency varies by up to ~200× across configurations; BPE-based splitting is needlessly slow, and BM25 + word splitting offers the best quality–latency trade-off. Thus, we provide evidence-based recommendations for implementing effective code-oriented RAG systems based on task requirements, model constraints, and computational efficiency.
\end{abstract}

\section{Introduction}

Large language models (LLMs) have rapidly advanced the state of the art in code intelligence, demonstrating impressive capabilities in code synthesis, refactoring, and defect detection –– combining both generation and retrieval‐based enhancements~\citep{RAPGen2023, evor2024, tan2025, zhou2025}. Despite these advances, even the largest models struggle to internalize the long-tail knowledge scattered throughout real-world software projects~\citep{LCA, LongTail2023, LongTailImpact2024}. Retrieval-augmented generation (RAG) tackles this limitation by augmenting the model’s prompt with text retrieved on-the-fly from an external corpus ~\citep{lewis2020retrieval}.

While RAG has become standard for open-domain question answering, its adaptation to software engineering tasks remains insufficiently examined. Source code presents unique challenges compared to natural language: it is highly structured, often exceeds typical context windows, and combines multiple languages and modalities (e.g., code, comments, issue reports). As a result, text-based RAG best practices do not transfer directly to code. Prior work has addressed individual aspects of this domain. Sparse lexical methods such as BM25~\citep{okapi1997, robertson2009bm25} remain strong baselines for code search~\citep{repocoder, zhou2023docprompting}, while dense models like CodeBERT~\cite{feng2020codebert} and E5~\cite{wang2022multilingual} offer improved cross-modal alignment. More specialized retrieval approaches have explored leveraging the semantic and syntactic structure of code, including syntax-aware chunking strategies~\citep{evor2024, zhou2025} and the use of repository-specific dependency graphs~\citep{draco, cAST2025} to guide retrieval. Yet, to the best of our knowledge, no study has systematically compared these design choices under a unified RAG pipeline across multiple code-centric tasks.

In this paper, we fill that gap by benchmarking a spectrum of RAG configurations on two complementary tasks from the Long Code Arena (LCA) benchmark collection~\cite{LCA}:

\textbf{Code completion (CC)}. The task is to generate the next line of code, conditioned on the preceding code context. This task involves programming language to programming language (PL$\rightarrow$PL) retrieval.

\textbf{Bug localization (BL)}  Identifying target files required correction based on issue text. This task involves natural language to programming language (NL$\rightarrow$PL) retrieval.

Both tasks require accurate retrieval of semantically related and often distant context snippets before generation. We then use these tasks to explore mixtures of various approaches coming from three orthogonal RAG axes: the \textbf{chunking strategy} --- dividing code into whole files, fixed-sized segments, or syntax-aware chunks; the \textbf{similarity scorer} --- sparse metrics such as BM25 or IoU, dense encoders, and repository structure-aware approaches; and the \textbf{tokenization granularity} (for sparse retrieval only) --- measuring similarity at the level of BPE tokens, words, or lines of code. By jointly varying these dimensions, we aim to clarify their individual and combined contributions to retrieval effectiveness in code-focused RAG. In addition, our experimental design varies the model context window from 128 to 16,000 tokens to study interactions between retrieval granularity and LLM capacity. The study yields several actionable insights.

We find that there is no single retrieval configuration that excels at all code tasks --- each requires its own optimal approach. For code completion (PL$\rightarrow$PL), lightweight, sparse retrieval with BM25 and word-level splitting stands out, significantly outperforming more complex dense models both in accuracy and speed. In contrast, bug localization (NL$\rightarrow$PL) benefits far more from advanced dense embeddings, which provide much stronger alignment between natural language and code than traditional sparse methods. We demonstrate that latency varies by up to 200× between configurations, emphasizing the importance of holistic cost-quality trade-offs for practical deployment.

Chunk size also matters: smaller code chunks (32-64 lines) are best for models with limited context windows ($\le$4,000 tokens), while whole-file retrieval becomes effective as the available context grows (up to 16,000 tokens). Surprisingly, simple line-based chunking is consistently as effective as syntax-aware approaches, suggesting that elaborate code parsing offers limited benefit.

Our findings serve as evidence-based guidelines for practitioners who must choose between accuracy, speed, and engineering complexity when building RAG systems for software engineering.

\section{Related Work}

While RAG is well studied for open-domain QA, its application to software engineering tasks remains relatively underexplored~\cite{ragReviewOche2025}. Recent work in retrieval for code and software engineering tasks demonstrates a rich diversity of approaches, encompassing lexical, dense, hybrid, and structure-aware techniques, each balancing quality and latency in distinct ways.

\textbf{Dense and sparse similarity scorers} remain competitive in code tasks ~\citep{robertson2009bm25, repocoder,zhou2023docprompting}, while dense models such as CodeBERT~\cite{feng2020codebert}, GraphCodeBERT~\cite{guo2020graphcodebert}, UniXcoder~\cite{guo2022unixcoder}, and E5~\cite{wang2022multilingual} improve cross-modal alignment. Hybrid search~\citep{lin2021framework} and learned sparse models like SPLADE~\cite{splade} aim to combine precision, recall, and efficiency. While these approaches have highlighted different strengths depending on the retrieval scenario, prior work generally selects one retrieval paradigm for a specific setting, lacking systematic benchmarking of retrieval strategies across multiple tasks and under various computational constraints.

\textbf{Chunking strategies} vary from line-based windows, offering language-agnostic simplicity, to syntax-aware splitting that preserves structural coherence~\citep{evor2024,zhou2025}, and graph-guided retrieval using dependency or call graphs~\citep{coderag,draco}. While some studies show semantic chunking can yield performance gains~\citep{cAST2025, ChunkRAG, denseX}, others argue that the computational costs associated with semantic chunking are not justified by consistent performance gains ~\cite{semanticchunk}. Further, works like~\cite{HierarcChunk} explore optimal chunk sizes but are limited in scope. Despite these efforts, it remains unclear how chunking choices interact with retrieval methods and task modalities in realistic code-focused RAG pipelines.

% \textbf{Tokenization granularity} for sparse retrieval --- such as the choice between token-level, word-level, or line-level splitting --- is typically treated as a low-level implementation detail rather than a focus of systematic investigation. Most code retrieval studies do not explicitly compare or report on the impact of different granularities within sparse retrieval pipelines, despite the potential implications for both efficiency and retrieval quality. By explicitly evaluating tokenization granularity as a separate axis in our experiments, we address this overlooked aspect and provide a targeted analysis of its effects on both latency and accuracy in code-focused RAG.

% \textbf{Tokenization granularity} for sparse retrieval is typically treated as a low-level implementation detail rather than a focus of systematic investigation and scarcely analyzed. ~\cite{SplitterSparse2021} benchmarked subword, token and char splitting for BM25 retrieval on the cross-lingual retrieval. \cite{SplitterSparseRareLangs2025} benchmarked word-level, token-level and language-specific tokenizer on multilingial QA task. Both of these works found that token-level text splitting on the considered tasks provide much better performance.

\textbf{Tokenization granularity} for sparse retrieval --- as in token-, word-, or line-level splitting --- is seldom studied systematically and is often treated as an implementation detail. While some recent work has compared some granularities for natural language tasks~\cite{SplitterSparse2021,SplitterSparseRareLangs2025} and found token-level splitting performs best, such findings have not been extended to code-focused RAG, nor have efficiency implications been explored. We address this gap by explicitly evaluating tokenization granularity as a separate axis in our experiments.

\textbf{Efficiency–quality trade-offs} have been the focus of recent studies, which highlight challenges in system latency and resource consumption. Recent works, such as ~\citep{RAGLatencytradeoff, METISFastRag}, provide a thorough analysis, mainly focusing on indexing techniques for dense retrieval. Our work complements these directions by systematically benchmarking retrieval design choices for code-related RAG tasks, extending the analysis to a broader range of retrieval techniques.

\textbf{Task-specific pipelines}, such as ReACC, RepoFormer and ProCC ~\citep{reacc, repoformer, ProCC} for retrieval-augmented code completion, ReCo~\cite{reco} for stylistic normalization in code search, and CodeRAG~\cite{coderag}, which leverages requirement–code graphs, demonstrate substantial gains on their respective tasks. These targeted solutions illustrate that retrieval choices --- including representation, chunking, and the use of code structure --- can have significant task-dependent impacts.

Collectively, these works highlight the versatility of retrieval approaches --- lexical, dense, hybrid, and structure-aware. Each exhibiting distinct quality --- latency trade-offs that map onto retrieval design axes such as chunking strategy, similarity scoring, and tokenization granularity. However, most prior studies assess these factors in isolation or optimize for a single configuration, leaving open the question of how these choices interact under realistic compute constraints and varying task modalities. Our work addresses this gap by benchmarking diverse retrieval configurations across two complementary code tasks under varying context windows and latency constraints, enabling evidence-based recommendations for task- and budget-aware code-oriented RAG system design.

\section{Experimental Setup}
\label{sec:setup}

\subsection{Tasks and Metrics}
\label{subsec:tasks}

\textbf{Dataset and scope.}  All experiments are run on the Long Code Arena (LCA) benchmark~\cite{LCA}. Retrieval is restricted to the target repository. We do not train or fine-tune any models.

\textbf{Code Completion (CC).}  
The task is to generate the next line of code, conditioned on the preceding code context. Target lines are selected to reference code entities --- classes or methods --- defined in other files of the same repository. Primary retrieval metric is an exact match (EM): we record EM = 1 if the target line matches the generated one; otherwise, EM = 0. We report the mean EM over instances. The retrieval task is to retrieve the code according to the context of the completion file of the chunk (PL-PL), without seeing the target line itself. Generation is then performed with \textsc{DeepSeek-Coder-1.3B}. The tasks are provided for two languages: Kotlin and Python.

\textbf{Bug Localization (BL).}  
Given a natural-language issue description, the retrieval task is to rank repository files by likelihood of containing the described bug. The ground truth set is unordered and may contain several files. Following the LCA work, we evaluate ranking with Normalized Discounted Cumulative Gain (NDCG)~\cite{ndcg}. The tasks are provided for three languages: Java, Kotlin, and Python.

\subsection{Retrieval Components}
\label{subsec:components}

We factor retrieval into three orthogonal sub-modules.

    \textbf{Chunker} maps each source file to a set of textual chunks to be indexed. We consider (i) whole files, a single chunk per file; (ii) fixed-length non-overlapping windows of 8, 16, 32, 64, or 128 lines, and (iii) syntax-aware recursive splitter via \textsc{langchain}~\cite{langchain} with a target length matched to the corresponding fixed-line alternative.
         
    \textbf{Splitter} (applies for sparse retriever only) splits a chunk into a bag of "tokens" consumed by the scorer. Options are (i) \emph{line-level}, each distinct line is a token, (ii) \emph{word-level}, split on non-alphanumeric boundaries; drop punctuation and numerals; no stemming, and (iii) \emph{BPE}, byte-pair tokens.
    
    \textbf{Scorer} assigns a similarity score between query and chunk. We test (i) \emph{sparse} lexical scorers: IoU (normalized set overlap) and BM25 (Okapi), (ii) \emph{dense} bi-encoder models: E5-small/base/large~\cite{wang2022multilingual}, Voyager-3/Code/Lite~\cite{voyageai}, (iii) \emph{Structure-aware}: DraCo (dataflow/dependency graph) and a path-distance heuristic. \emph{DraCo} (Dataflow-guided Retrieval Augmentation) ~\cite{draco} leverages static dataflow analysis to collect the context. \emph{Path-distance} heuristic retrieves context based on how close files are in the directory tree, assuming that files located nearer to each other are more likely to be relevant.

The retrieval configuration is therefore a triplet $(chunker, splitter, scorer)$.  
Unless noted, dense encoders use a 512-token limit; longer documents are truncated from right.

\subsection{Packing and budgets}

\textbf{Query construction} CC: the query is the last chunk from the completion file with the target line removed. BL: the query is the issue text.

\textbf{Ranking and packing} For each query, we rank candidate chunks/files. We then greedily pack the top-ranked items into the prompt until a token budget is reached, preserving rank order and discarding overflows. Budgets for CC are 128, 4,096, 8,192, and 16,384 tokens. BL is evaluated as a ranking task over files (NDCG) and does not use a packing budget.

\textbf{Hybrids and reranking} Structure-aware lists (DraCo) are combined with the best-performing sparse/dense retriever by prepending the structure-aware candidates and backfilling with BM25-ranked items to fill the budget. If the structure-aware method returns fewer candidates than needed, we continue with BM25 order.

\subsection{Hyper-parameter Search}

A full grid over all axes would require an excessive number of runs, so we adopt a staged search:

\begin{enumerate}[label=Stage~\arabic*]
    \item \textbf{scorer $\!\times\!$ splitter.}  
          Using the whole file as a chunk, we benchmark every $(scorer, splitter)$ pair and select the best per task.  
          Results select BM25 + word for CC; E5-large and Voyger-3 for BL.
    \item \textbf{chunk size.}  
          Fixing the chosen at Step 1 scorer and splitter, we sweep chunk line windows $\in\{8,16,32,64,128\}$ and whole files. We measure EM (@CC) for each candidate at four LLM context budgets: $128, 4\,096, 8\,192, 16\,384$ tokens.
    \item \textbf{chunker type}  
          Compare the best fixed-line window from Stage 2 to syntax-aware recursive splitting with matched average lengths.
    \item \textbf{hybrid rerankers.}  
          Evaluate dataflow graph-based DraCo and path distance combined with the best sparse/dense retriever.
\end{enumerate}

The code completion task is sensitive to context size, so we report scores across a range of context lengths (128–16,000 tokens) and analyze how performance depends on context size and other factors. This progressive refinement methodology allowed us to systematically narrow the search space while identifying the most effective configuration components at each stage of the optimization process.

\subsection{Implementation notes}

Source files are indexed as-is (comments and strings retained). Word-level splitting removes punctuation and numerals; no stemming or identifier de-camelization is applied. We use dense encoders off-the-shelf. All latency measurements reported later are wall-clock times.

\section{Results and Analysis}

\subsection{Code Completion Task}

\begin{figure}[t]
    \centering
    \includegraphics[width=0.9\linewidth]{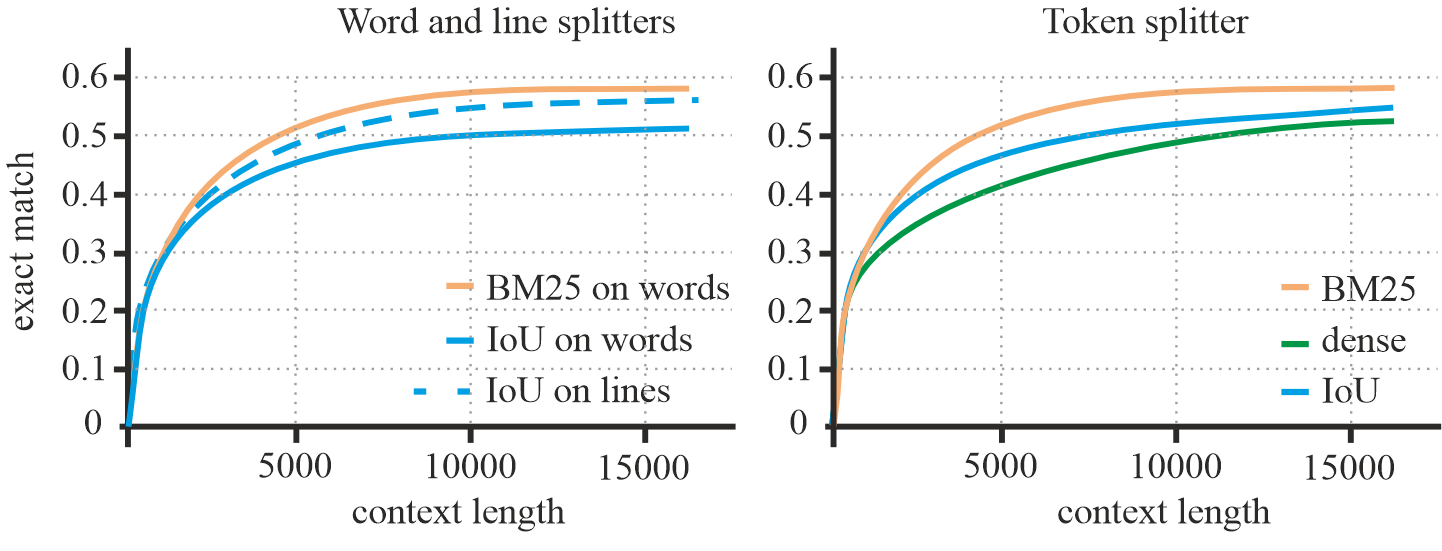}
    \caption{Exact match scores for various combinations of splitters and scorers.}
    \label{fig:split-scorer-cc}
\end{figure}

\subsubsection{Optimal Scorers and Splitters}

In this experiment, we fix the chunker to whole-file retrieval and sweep scorer–splitter pairs across context budgets. For sparse scorers (BM25, IoU), we evaluate line-, word-, and BPE-level splitters. Dense encoder (E5-large) operates on raw text and ignores the splitter choice. Retrieved files are packed greedily in rank order up to the context budget.
Figure \ref{fig:split-scorer-cc} summarizes the results for the evaluated splitter–scorer pairs. Three consistent trends emerge:

\textbf{BM25 dominates in accuracy.}  BM25 achieves the highest exact-match rate, outperforming both IoU and dense embeddings by $\approx 10$ p.p. on average for medium and large context lengths ($>2\,000$), where lexical overlap remains informative.

\textbf{Word splitter is more efficient than BPE, while showing same quality.}  Using BPE tokens or plain words with BM25 yields statistically indistinguishable EM; however, the word splitter is 9× faster to build. We therefore adopt the BM25 scorer with word splitter as our default sparse configuration.

\textbf{Dense encoders plateau early.}  E5-large trails BM25, making sparse methods the pragmatic choice for PL→PL retrieval, particularly for longer documents, which require more resources to encode. 

\textbf{IoU with line splitting} sits in the middle ground, cheaper than BM25 but also $\approx 4$ p.p. worse on EM, making it attractive only for latency-critical deployments.

\subsubsection{Optimal Chunking}

We study chunk-size choice under the best sparse configuration from the previous subsection: BM25 with word-level splitting. We vary two knobs:

\textbf{Index chunk size} $L_i$ $\{8, 16, 32, 64, 128, \inf\}$. For every file, including the completion file, we index non-overlapping windows of a $L_i$ lines; $\inf$ denotes whole-file indexing.

\textbf{Query window size} $L_q$ $\{8, 16, 32, 64, 128\}$. For completion file, query consists of the last $L_q$ lines preceding the target line.
Retrieved chunks are greedily packed in rank order up to the budget tokens.

\begin{figure}
    \centering
    \includegraphics[width=0.9\linewidth]{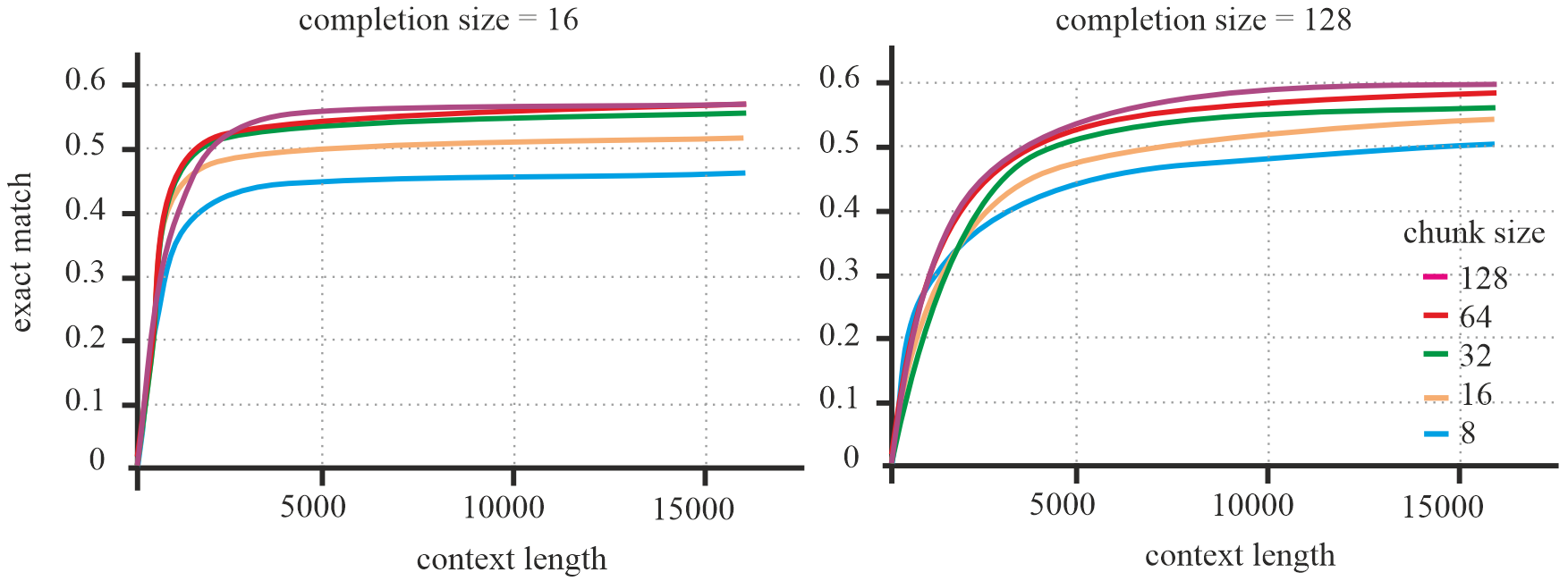}
    \caption{Exact match scores for various completion file sizes and chunk size}
    \label{fig:chunk_sizes_cc}
\end{figure}

The results (Figure~\ref{fig:chunk_sizes_cc}) revealed a clear relationship between optimal chunk size and the model's context window length. Small chunks (8-16 lines) consistently underperformed across all context lengths, likely because they contained insufficient information to establish meaningful relevance. For models with shorter context windows (less than 4,000 tokens), moderate chunks of 32-64 lines provide the best performance and saturate faster. For models with longer context windows (> 8,000 tokens), larger completion chunk sizes of 128 lines or more yielded superior results at large context, but saturate more slowly with context size. Ultimately, when models had access to extensive context windows (16,000 tokens), the whole file approach performed on par with chunked alternatives. This suggests that smaller contexts benefit from more granular retrieval units that can precisely match the query, while longer contexts benefit from having more comprehensive code segments that preserve broader structural and contextual information.

Line splitting demonstrated a slight but consistent performance advantage over code structure-aware splitting across all parameters. This surprising result suggests that simple line-based chunking may preserve sufficient code coherence.

These findings provide valuable guidance for RAG system implementation. They suggest that chunk sizes should be dynamically adjusted based on the underlying generation model's context window length rather than using a one-size-fits-all approach. Additionally, results demonstrate that line splitting performs on par with a more sophisticated structure-aware chunking approach. It offers implementation simplicity and a language-agnostic application, making it a robust choice for general code retrieval systems.

\subsubsection{Structure-Aware Code Retrieval}

We evaluated the dataflow-guided DraCo retriever, leveraging the import dependency graph structure to identify files directly relevant to the completion target file as well as path-distance retriever. We augment DraCo's selections with additional files identified through BM25 ranking, as DraCo returns only limited number of files. 

\begin{figure}
    \centering
    \includegraphics[width=0.5\linewidth]{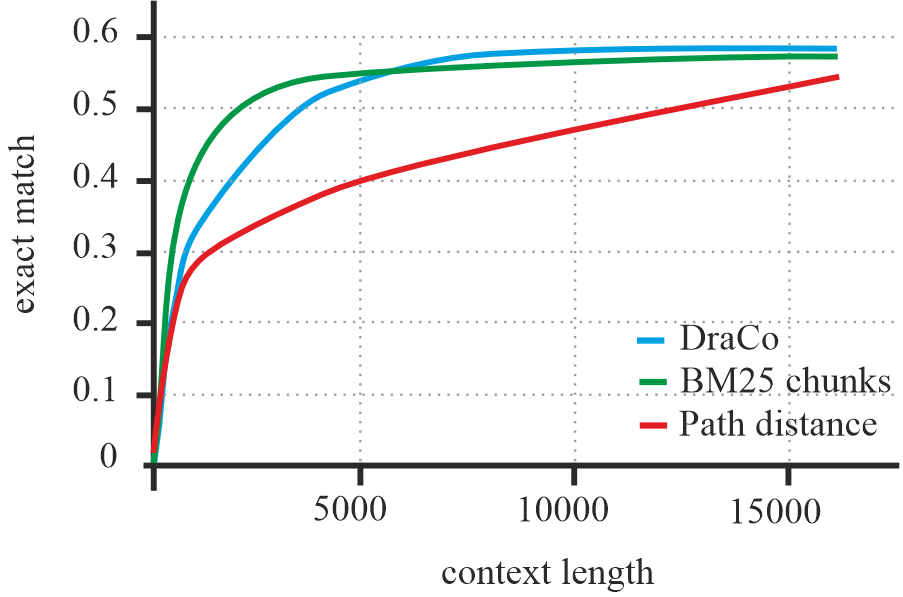}
    \caption{Exact match scores of benchmarked retrievers.}
    \label{fig:composers}
\end{figure}

The results (Figure~\ref{fig:composers}) indicate that chunking-based retrieval provides superior accuracy for medium contexts (<~8,000 tokens). However, as context length increases, the DraCo and chunking approaches converge towards similar performance levels. Path distance retrieval remains consistently lower-performing, suggesting that directory structure alone is insufficient for optimal context selection for the code completion task.

\subsection{Bug Localization Task}

We evaluate bug localization as NL→PL file ranking with NDCG. Sparse methods (BM25, IoU) use line-, word-, or BPE-level splitting. Dense encoders (E5 small/base/large and Voyager-3 family) operate on raw text with model-specific input limits; overlength files were truncated from the right.

\begin{figure}
    \centering
    \includegraphics[width=1\linewidth]{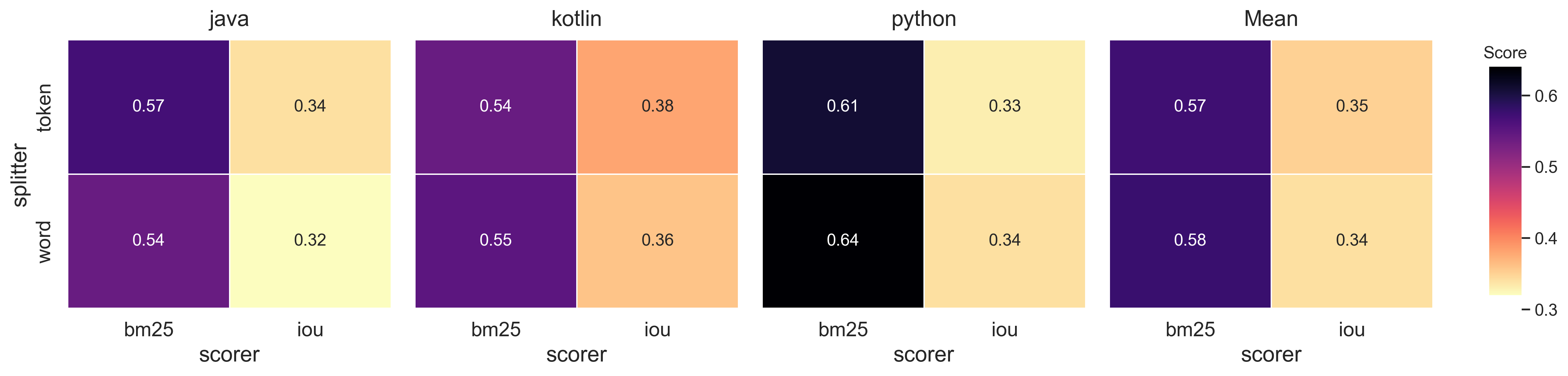}
    \caption{NDCG scores for various combinations of splitters and scorers for sparse retrieval}
    \label{fig:split-scorer-bug-loc}
\end{figure}

Figure~\ref{fig:split-scorer-bug-loc} summarizes sparse variants. BM25 consistently outperforms IoU across Java, Kotlin, and Python, reflecting the limits of exact-overlap signals in cross-modal queries. Table~\ref{tab:ndcg_models} extends the comparison to dense encoders and shows that dense retrieval is stronger on average: E5-large reaches a mean NDCG of 0.59 versus 0.57 for BM25 with a word splitter, despite the 512-token limit. Proprietary Voyager embeddings are substantially higher, with Voyager-3-Code at 0.72 mean NDCG. Language-wise, the gap narrows for Python: BM25+word attains 0.64 versus 0.61 for E5-large, suggesting that repositories with richer natural-language artifacts in code and comments reduce the advantage of semantic encoders.

Efficiency varies greatly with splitting strategy. BM25 with a word splitter is the fastest (0.07 s per 1M symbols), while BM25 with BPE/token splitting is ~16× slower (1.15 s) without quality gains and can even lag behind lighter dense encoders (E5-small/base) due to larger context size. E5-large requires 2.8 s, while Voyager-3-Code (512) is substantially slower at 19 s; the 32000-token Voyager variants are even slower due to technical token-budget constraints at the time of experiments. These findings show that, for NL→PL ranking on large repositories (here roughly 20\% of files exceed 15000 tokens), dense models offer better accuracy even with truncation, but sparse BM25 with word-level splitting remains an excellent latency–quality baseline. When accuracy is paramount and cost permits, Voyager-3-Code leads by a wide margin; when latency or budget is critical, BM25+word is the pragmatic choice.

\begin{table}
\centering
\caption{NDCG scores for various models on the bug localization task.}
\label{tab:ndcg_models}
\begin{tabular}{@{}lccccccc@{}}
\toprule
\textbf{Model} & \textbf{\shortstack{Embed\\dim}} & \textbf{\shortstack{Context\\length}} & \textbf{\shortstack{Time,\\s/1M symb*}} & \multicolumn{4}{c}{\textbf{NDCG}} \\
\cmidrule(l){5-8}
 &  &  &  & \textbf{java} & \textbf{kotlin} & \textbf{python} & \textbf{mean} \\
\midrule
Voyager-3-code 512 & 1,024 & 512 & 19 & \textbf{0.712} & \textbf{0.701} & \textbf{0.738} & \textbf{0.717} \\
Voyager-3-code & 1,024 & 32,000 & 110** & 0.708 & 0.676 & 0.705 & 0.696 \\
Voyager-3 & 1,024 & 32,000 & 32 & 0.660 & 0.697 & 0.651 & 0.670 \\
Voyager-3-lite & 512 & 32,000 & 17 & 0.647 & 0.637 & 0.625 & 0.636 \\
BM25, word splitter & — & $\infty$ & \textbf{0.07} & 0.541 & 0.547 & 0.635 & 0.574 \\
BM25, token splitter & — & $\infty$ & 1.2 & 0.573 & 0.538 & 0.614 & 0.575 \\
E5 large 560M & 1,024 & 514 & 2.8 & 0.596 & 0.569 & 0.606 & 0.590 \\
E5-base 278M & 768 & 514 & 1.1 & 0.446 & 0.478 & 0.467 & 0.464 \\
E5-small 117M & 384 & 514 & 0.7 & 0.409 & 0.511 & 0.491 & 0.470 \\
\bottomrule
\end{tabular}
\begin{flushleft}
\small\textit{*retrieval time includes encoding the whole repository for each data point. Symbols are counted all - even truncated ones, which were not used in the encoder.}

\small\textit{**retrieval time for Voyager-3-code is exceedingly high because we have to set batch\_size=2 due to the limit of 120,000 tokens per request.}

\end{flushleft}
\end{table}

\subsection{Latency Analysis}

We assess end-to-end retrieval latency normalized by repository size, including query tokenization, splitting and similarity scoring. For sparse methods, we use prebuilt inverted indices; for dense encoders, we report query-time document encoding plus similarity computation, which makes the dense figures an upper bound for deployments that could precompute embeddings.

Table~\ref{tab:latency} summarizes results on the code completion setup. The path-distance heuristic is effectively free because it ignores content and operates only on directory metadata. Among content-based methods, IoU with line-level splitting is the fastest at 0.02 s/1M symbols; moving to word-level splitting raises cost to 0.1 s due to a larger token inventory per document. BM25 with word-level splitting remains reasonably efficient at 0.2 s and, as shown earlier, offers the best quality–latency balance for PL→PL. BPE/token-level pipelines are an order of magnitude slower (IoU 1.5 s; BM25 2.0 s) without accuracy gains, reflecting tokenizer overhead and much larger posting lists. Dense retrieval with E5-large is the slowest at 3 s/1M symbols because it requires a forward pass over all candidate chunks.

For a representative repository of 2.3 million symbols, these rates translate to approximately 1.2 ms for path distance, 0.04 s for IoU-line, 0.5 s for BM25-word, 3.5 s for IoU-token, 4.5 s for BM25-token, and 7.5 s for E5-large. Thus, the spread between the fastest practical content-based scorer (IoU-line) and a dense encoder is about 180×, and between BM25-word and E5-large about 14×, in line with the “orders of magnitude” difference noted earlier.

The bug localization task presents a tougher efficiency–effectiveness trade-off. We recommend BM25-based retrieval over E5-large: it offers similar performance but is about 40× faster. Proprietary Voyage models achieve higher scores, but with much greater latency and cost. Ultimately, the choice is yours: use BM25 if speed and cost are paramount, or choose Voyage if you prioritize accuracy despite higher latency and expense.

\begin{table}
\centering
\caption{Retrieval latency and EM scores (the best setup) for different approaches for the code completion task.}
\label{tab:latency}
\begin{tabular}{@{}lccc@{}}
\toprule
\textbf{Scorer} & \textbf{Splitter} & \textbf{Time, s/1M repo symbols} & \textbf{EM at 4K / 16K context}  \\
\midrule
Path distance & -- & 0.0005 & 0.37 / 0.55 \\
IoU & line & 0.02 & 0.46 / 0.57 \\
IoU & word & 0.1 & 0.44 / 0.51 \\
BM25 & word & 0.2 & \textbf{0.55 / 0.60} \\
IoU & token & 1.5 & 0.40 / 0.55 \\
BM25 & token & 2.0 & 0.50 / 0.59 \\
Dense (E5-large, 512) & -- & 3.3 & 0.39 / 0.52 \\
\bottomrule
\end{tabular}

\begin{flushleft}
\end{flushleft}
\end{table}

\section{Limitations and Future Work}

For the retrieval part, we have not tested deduplication techniques to remove redundant information from retrieved contexts or context compression methods to maximize the utility of limited context windows. This could transform context window constraints, enabling even small-context models to effectively process information from larger code segments. However, context compression will come at some additional latency cost.

This study focuses on two LCA tasks — single-line code completion (PL→PL) and bug localization (NL→PL) — across Java, Kotlin, and Python with repository-local retrieval. While these tasks are representative, the scope limits external validity: results may differ for multi-line completion/repair or long-form generation. We evaluated generation with a single small model (DeepSeek-Coder-1.3B) to isolate retrieval effects; preliminary checks with a larger variant showed similar trends. We anticipate that the core principles would remain consistent with our findings on this narrow set of tasks.

Integrating additional code-specific structural information could enhance retrieval effectiveness, particularly for more complex tasks than code completion.

\section{Conclusions and Recommendations}

We systematically evaluated retrieval design choices for code-oriented RAG on two complementary tasks -- PL→PL code completion and NL→PL bug localization. By varying chunking, similarity scoring, and splitting granularity across different context budgets, we provide critical insights to guide the implementation of effective RAG systems for software engineering. 

\textbf{Sparse vs Dense}. For code completion tasks involving PL-PL retrieval, sparse retrieval methods — particularly BM25 with word-level splitting — consistently delivered superior results with reasonable computational efficiency. In contrast, bug localization tasks that involve NL-PL retrieval showed markedly different behaviour. Dense retrieval methods using neural embeddings mostly outperformed sparse retrieval approaches, with models like E5-large marginally higher NDCG scores than BM25, and Voyager-3-Code, achieving $\approx 0.72$ NDCG score compared to $\approx 0.57$ achieved by BM25.

These results reflect a modality-driven divide. In PL→PL code completion, substantial lexical overlap between the query and target makes sparse lexical scoring (e.g., BM25 with word-level splitting) a natural fit. In contrast, NL→PL bug localization benefits from dense encoders that capture semantic correspondence, bridging the gap between natural-language issue text and code implementations.

\textbf{Chunking}. There is an important relationship between optimal chunk size and available context window. Moderate windows (32–64 lines) are best for small budgets ($\leq 4000$ tokens), larger windows (64–128 lines) help as budgets grow, and whole-file retrieval becomes competitive at 16000 tokens. This scaling relationship underscores the importance of aligning retrieval granularity with model capacity — smaller contexts benefit from precise, focused chunks that maximize the relevance density, while larger contexts leverage broader context to better understand complex code relationships.

\textbf{Sophisticated chunking}. Line-based chunking matched or slightly exceeded syntax-aware splitting across budgets. We attribute this result to the fact that code completion relies more heavily on identifying semantically similar fragments rather than hierarchically related parent code structures in the codebase. This suggests that syntactic structure preservation may not be as critical as previously assumed for effective code retrieval, at least for the tasks evaluated in this study. This finding aligns with similar results reported for natural language tasks~\cite{semanticchunk}.

\textbf{Latency}. From a practical implementation perspective, our latency analysis revealed dramatic efficiency differences between retrieval approaches, with up to 180× speed differences between the fastest meaningful (IoU-line) and slowest (dense) methods. This emphasize the importance of balancing retrieval quality with computational efficiency, particularly for interactive applications where response time is critical.

Based on these results, we offer the following recommendations for code RAG systems:

\begin{enumerate}

    \item Match the retrieval strategy to the task nature. Use BM25 with word splitting for code-to-code retrieval tasks and dense embedding methods for natural language-to-code retrieval scenarios. The additional computational cost of dense retrieval is justified by its superior performance in cross-modal scenarios.

    \item Align chunk granularity with model context capacity. As model context windows expand, increase chunk sizes accordingly. For models with shorter context window ($\leq4000$ tokens), use 32-64 line chunks; for medium-length models (4000-8000 tokens), use 64-128 line chunks; and for models with long context windows ($\geq16000$ tokens), consider retrieving entire files when appropriate. This applies to both index chunk size and query windowing.

    \item Prioritize implementation efficiency through strategic choices. The word splitter offers nearly identical performance to tokenizer-based approaches at a fraction of the computational cost. For applications with strict latency requirements, consider using IoU with line splitting, which delivers reasonable performance with minimal computational overhead.

    \item Complexity does not necessarily improve quality. As we show for the code completion task, structural information did not enhance performance. However, for bug localization, including file paths and import statements in the context significantly improved performance.

    \item Consider the entire retrieval pipeline holistically. The most effective RAG systems will likely combine multiple retrieval strategies with the specific configuration tailored to the task requirements, codebase characteristics, and computational constraints.

\end{enumerate}

\section{Acknowledgments}

The authors thank Dr. Yaroslav Zharov for his valuable suggestions regarding the writing of the manuscript.

\bibliographystyle{plainnat}
\bibliography{paper}

\begin{thebibliography}{37}
\providecommand{\natexlab}[1]{#1}
\providecommand{\url}[1]{\texttt{#1}}
\expandafter\ifx\csname urlstyle\endcsname\relax
  \providecommand{\doi}[1]{doi: #1}\else
  \providecommand{\doi}{doi: \begingroup \urlstyle{rm}\Url}\fi

\bibitem[Beaulieu et~al.(1997)Beaulieu, Gatford, Huang, Robertson, Walker, and Williams]{okapi1997}
Micheline Beaulieu, Mike Gatford, Xiangji Huang, Stephen Robertson, Steve Walker, and P~Williams.
\newblock Okapi at trec-5.
\newblock \emph{Nist Special Publication SP}, pages 143--166, 1997.

\bibitem[Bogomolov et~al.(2024)Bogomolov, Eliseeva, Galimzyanov, Glukhov, Shapkin, Tigina, Golubev, Kovrigin, van Deursen, Izadi, and Bryksin]{LCA}
Egor Bogomolov, Aleksandra Eliseeva, Timur Galimzyanov, Evgeniy Glukhov, Anton Shapkin, Maria Tigina, Yaroslav Golubev, Alexander Kovrigin, Arie van Deursen, Maliheh Izadi, and Timofey Bryksin.
\newblock Long code arena: a set of benchmarks for long-context code models, 2024.
\newblock URL \url{https://arxiv.org/abs/2406.11612}.

\bibitem[Chase and Contributors(2022)]{langchain}
Harrison Chase and LangChain Contributors.
\newblock Langchain.
\newblock \url{https://github.com/langchain-ai/langchain}, 2022.
\newblock Accessed: 2025-08-19.

\bibitem[Chen et~al.(2024)Chen, Wang, Chen, Yu, Ma, Zhao, Zhang, and Yu]{denseX}
Tong Chen, Hongwei Wang, Sihao Chen, Wenhao Yu, Kaixin Ma, Xinran Zhao, Hongming Zhang, and Dong Yu.
\newblock Dense {X} retrieval: What retrieval granularity should we use?
\newblock In Yaser Al-Onaizan, Mohit Bansal, and Yun-Nung Chen, editors, \emph{Proceedings of the 2024 Conference on Empirical Methods in Natural Language Processing}, pages 15159--15177, Miami, Florida, USA, November 2024. Association for Computational Linguistics.
\newblock \doi{10.18653/v1/2024.emnlp-main.845}.
\newblock URL \url{https://aclanthology.org/2024.emnlp-main.845/}.

\bibitem[Cheng et~al.(2024)Cheng, Wu, and Hu]{draco}
Wei Cheng, Yuhan Wu, and Wei Hu.
\newblock Dataflow-guided retrieval augmentation for repository-level code completion.
\newblock In Lun-Wei Ku, Andre Martins, and Vivek Srikumar, editors, \emph{Proceedings of the 62nd Annual Meeting of the Association for Computational Linguistics (Volume 1: Long Papers)}, pages 7957--7977, Bangkok, Thailand, August 2024. Association for Computational Linguistics.
\newblock \doi{10.18653/v1/2024.acl-long.431}.
\newblock URL \url{https://aclanthology.org/2024.acl-long.431/}.

\bibitem[Feng et~al.(2020)Feng, Guo, Tang, and et~al.]{feng2020codebert}
Zhangyin Feng, Daya Guo, Duyu Tang, and et~al.
\newblock {CodeBERT}: A pre-trained model for programming and natural languages.
\newblock In \emph{Findings of EMNLP}, pages 1536--1547, 2020.

\bibitem[Formal et~al.(2021)Formal, Lassance, Piwowarski, and Clinchant]{splade}
Thibault Formal, Carlos Lassance, Benjamin Piwowarski, and Stéphane Clinchant.
\newblock Splade v2: Sparse lexical and expansion model for information retrieval, 2021.
\newblock URL \url{https://arxiv.org/abs/2109.10086}.

\bibitem[Guo et~al.(2020)Guo, Ren, Lu, and et~al.]{guo2020graphcodebert}
Daya Guo, Shuo Ren, Shuai Lu, and et~al.
\newblock {GraphCodeBERT}: Pre-training code representations with data flow.
\newblock arXiv preprint arXiv:2009.08366, 2020.

\bibitem[Guo et~al.(2022)Guo, Ren, Lu, and et~al.]{guo2022unixcoder}
Daya Guo, Shuo Ren, Shuai Lu, and et~al.
\newblock {UniXcoder}: Unified cross-modal pre-training for code representation.
\newblock In \emph{Proceedings of EMNLP}, pages 2715--2725, 2022.

\bibitem[J\"{a}rvelin and Kek\"{a}l\"{a}inen(2002)]{ndcg}
Kalervo J\"{a}rvelin and Jaana Kek\"{a}l\"{a}inen.
\newblock Cumulated gain-based evaluation of ir techniques.
\newblock \emph{ACM Trans. Inf. Syst.}, 20\penalty0 (4):\penalty0 422–446, October 2002.
\newblock ISSN 1046-8188.
\newblock \doi{10.1145/582415.582418}.
\newblock URL \url{https://doi.org/10.1145/582415.582418}.

\bibitem[Kandpal et~al.(2023)Kandpal, Deng, Roberts, Wallace, and Raffel]{LongTail2023}
Nikhil Kandpal, Haikang Deng, Adam Roberts, Eric Wallace, and Colin Raffel.
\newblock Large language models struggle to learn long-tail knowledge.
\newblock In \emph{Proceedings of the 40th International Conference on Machine Learning}, ICML'23. JMLR.org, 2023.

\bibitem[Lewis et~al.(2020)Lewis, Perez, Piktus, and et~al.]{lewis2020retrieval}
Patrick Lewis, Ethan Perez, Aleksandra Piktus, and et~al.
\newblock Retrieval-augmented generation for knowledge-intensive {NLP} tasks.
\newblock In \emph{Advances in Neural Information Processing Systems (NeurIPS)}, 2020.

\bibitem[Li et~al.(2024)Li, Zhou, and Shen]{reco}
Haochen Li, Xin Zhou, and Zhiqi Shen.
\newblock Rewriting the code: A simple method for large language model augmented code search.
\newblock In Lun-Wei Ku, Andre Martins, and Vivek Srikumar, editors, \emph{Proceedings of the 62nd Annual Meeting of the Association for Computational Linguistics (Volume 1: Long Papers)}, pages 1371--1389, Bangkok, Thailand, August 2024. Association for Computational Linguistics.
\newblock \doi{10.18653/v1/2024.acl-long.75}.
\newblock URL \url{https://aclanthology.org/2024.acl-long.75/}.

\bibitem[Li et~al.(2025)Li, Shi, Zhang, Li, Li, Tao, Li, Liu, Tao, and Jin]{coderag}
Jia Li, Xianjie Shi, Kechi Zhang, Lei Li, Ge~Li, Zhengwei Tao, Jia Li, Fang Liu, Chongyang Tao, and Zhi Jin.
\newblock Coderag: Supportive code retrieval on bigraph for real-world code generation, 2025.
\newblock URL \url{https://arxiv.org/abs/2504.10046}.

\bibitem[Lin(2022)]{lin2021framework}
Jimmy Lin.
\newblock A proposed conceptual framework for a representational approach to information retrieval.
\newblock \emph{SIGIR Forum}, 55\penalty0 (2), March 2022.
\newblock ISSN 0163-5840.
\newblock \doi{10.1145/3527546.3527552}.
\newblock URL \url{https://doi.org/10.1145/3527546.3527552}.

\bibitem[Lu et~al.(2022)Lu, Duan, Han, Guo, Hwang, and Svyatkovskiy]{reacc}
Shuai Lu, Nan Duan, Hojae Han, Daya Guo, Seung-won Hwang, and Alexey Svyatkovskiy.
\newblock {R}e{ACC}: A retrieval-augmented code completion framework.
\newblock In Smaranda Muresan, Preslav Nakov, and Aline Villavicencio, editors, \emph{Proceedings of the 60th Annual Meeting of the Association for Computational Linguistics (Volume 1: Long Papers)}, pages 6227--6240, Dublin, Ireland, May 2022. Association for Computational Linguistics.
\newblock \doi{10.18653/v1/2022.acl-long.431}.
\newblock URL \url{https://aclanthology.org/2022.acl-long.431/}.

\bibitem[Nguyen et~al.(2025)Nguyen, Nguyen, and Nguyen]{HierarcChunk}
Hai-Toan Nguyen, Tien-Dat Nguyen, and Viet-Ha Nguyen.
\newblock Enhancing retrieval augmented generation with hierarchical text segmentation chunking.
\newblock In Wray Buntine, Morten Fjeld, Truyen Tran, Minh-Triet Tran, Binh Huynh Thi~Thanh, and Takumi Miyoshi, editors, \emph{Information and Communication Technology}, pages 209--220, Singapore, 2025. Springer Nature Singapore.

\bibitem[Oche et~al.(2025)Oche, Folashade, Ghosal, and Biswas]{ragReviewOche2025}
Agada~Joseph Oche, Ademola~Glory Folashade, Tirthankar Ghosal, and Arpan Biswas.
\newblock A systematic review of key retrieval-augmented generation (rag) systems: Progress, gaps, and future directions, 2025.
\newblock URL \url{https://arxiv.org/abs/2507.18910}.

\bibitem[Ogundepo et~al.(2022)Ogundepo, Zhang, and Lin]{SplitterSparseRareLangs2025}
Odunayo Ogundepo, Xinyu Zhang, and Jimmy Lin.
\newblock Better than whitespace: Information retrieval for languages without custom tokenizers, 2022.
\newblock URL \url{https://arxiv.org/abs/2210.05481}.

\bibitem[Qu et~al.(2025)Qu, Tu, and Bao]{semanticchunk}
Renyi Qu, Ruixuan Tu, and Forrest~Sheng Bao.
\newblock Is semantic chunking worth the computational cost?
\newblock In Luis Chiruzzo, Alan Ritter, and Lu~Wang, editors, \emph{Findings of the Association for Computational Linguistics: NAACL 2025}, pages 2155--2177, Albuquerque, New Mexico, April 2025. Association for Computational Linguistics.
\newblock ISBN 979-8-89176-195-7.
\newblock \doi{10.18653/v1/2025.findings-naacl.114}.
\newblock URL \url{https://aclanthology.org/2025.findings-naacl.114/}.

\bibitem[Ray et~al.(2025)Ray, Pan, Gu, Du, Feng, Ananthanarayanan, Netravali, and Jiang]{METISFastRag}
Siddhant Ray, Rui Pan, Zhuohan Gu, Kuntai Du, Shaoting Feng, Ganesh Ananthanarayanan, Ravi Netravali, and Junchen Jiang.
\newblock Metis: Fast quality-aware rag systems with configuration adaptation, 2025.
\newblock URL \url{https://arxiv.org/abs/2412.10543}.

\bibitem[Robertson and Zaragoza(2009)]{robertson2009bm25}
Stephen Robertson and Hugo Zaragoza.
\newblock The probabilistic relevance framework: {BM25} and beyond.
\newblock \emph{Foundations and Trends in Information Retrieval}, 3\penalty0 (4):\penalty0 333--389, 2009.
\newblock \doi{10.1561/1500000019}.

\bibitem[Shen et~al.(2024)Shen, Umar, Maeng, Suh, and Gupta]{RAGLatencytradeoff}
Michael Shen, Muhammad Umar, Kiwan Maeng, G.~Edward Suh, and Udit Gupta.
\newblock Towards understanding systems trade-offs in retrieval-augmented generation model inference, 2024.
\newblock URL \url{https://arxiv.org/abs/2412.11854}.

\bibitem[Singh et~al.(2025)Singh, Aggarwal, Allahverdiyev, Taha, Akalin, Zhu, and O'Brien]{ChunkRAG}
Ishneet~Sukhvinder Singh, Ritvik Aggarwal, Ibrahim Allahverdiyev, Muhammad Taha, Aslihan Akalin, Kevin Zhu, and Sean O'Brien.
\newblock Chunkrag: Novel llm-chunk filtering method for rag systems, 2025.
\newblock URL \url{https://arxiv.org/abs/2410.19572}.

\bibitem[Su et~al.(2024)Su, Jiang, Lai, Wu, Shi, Liu, Liu, and Yu]{evor2024}
Hongjin Su, Shuyang Jiang, Yuhang Lai, Haoyuan Wu, Boao Shi, Che Liu, Qian Liu, and Tao Yu.
\newblock {E}vo{R}: Evolving retrieval for code generation.
\newblock In Yaser Al-Onaizan, Mohit Bansal, and Yun-Nung Chen, editors, \emph{Findings of the Association for Computational Linguistics: EMNLP 2024}, pages 2538--2554, Miami, Florida, USA, November 2024. Association for Computational Linguistics.
\newblock \doi{10.18653/v1/2024.findings-emnlp.143}.
\newblock URL \url{https://aclanthology.org/2024.findings-emnlp.143/}.

\bibitem[Tan et~al.(2025{\natexlab{a}})Tan, Luo, Jiang, Zhan, Li, Zhang, and Zhang]{ProCC}
Hanzhuo Tan, Qi~Luo, Ling Jiang, Zizheng Zhan, Jing Li, Haotian Zhang, and Yuqun Zhang.
\newblock Prompt-based code completion via multi-retrieval augmented generation.
\newblock \emph{ACM Trans. Softw. Eng. Methodol.}, March 2025{\natexlab{a}}.
\newblock ISSN 1049-331X.
\newblock \doi{10.1145/3725812}.
\newblock URL \url{https://doi.org/10.1145/3725812}.
\newblock Just Accepted.

\bibitem[Tan et~al.(2025{\natexlab{b}})Tan, Luo, Jiang, Zhan, Li, Zhang, and Zhang]{tan2025}
Hanzhuo Tan, Qi~Luo, Ling Jiang, Zizheng Zhan, Jing Li, Haotian Zhang, and Yuqun Zhang.
\newblock Prompt-based code completion via multi-retrieval augmented generation.
\newblock \emph{ACM Trans. Softw. Eng. Methodol.}, March 2025{\natexlab{b}}.
\newblock ISSN 1049-331X.
\newblock \doi{10.1145/3725812}.
\newblock URL \url{https://doi.org/10.1145/3725812}.
\newblock Just Accepted.

\bibitem[{Voyage AI}(2025)]{voyageai}
{Voyage AI}.
\newblock Voyage ai documentation.
\newblock \url{https://docs.voyageai.com/}, 2025.
\newblock Accessed: 2025-08-19.

\bibitem[Wang et~al.(2022)Wang, Liu, Sun, and et~al.]{wang2022multilingual}
Sheng Wang, Yichao Liu, Zhihong Sun, and et~al.
\newblock Text embeddings by weakly-supervised contrastive learning.
\newblock In \emph{Findings of ACL}, pages 5110--5122, 2022.

\bibitem[Wang et~al.(2023)Wang, Wang, Joty, and Hoi]{RAPGen2023}
Weishi Wang, Yue Wang, Shafiq Joty, and Steven~C.H. Hoi.
\newblock Rap-gen: Retrieval-augmented patch generation with codet5 for automatic program repair.
\newblock In \emph{Proceedings of the 31st ACM Joint European Software Engineering Conference and Symposium on the Foundations of Software Engineering}, ESEC/FSE 2023, page 146–158, New York, NY, USA, 2023. Association for Computing Machinery.
\newblock ISBN 9798400703270.
\newblock \doi{10.1145/3611643.3616256}.
\newblock URL \url{https://doi.org/10.1145/3611643.3616256}.

\bibitem[Wu et~al.(2024)Wu, Ahmad, Zhang, Ramanathan, and Ma]{repoformer}
Di~Wu, Wasi Ahmad, Dejiao Zhang, Murali~Krishna Ramanathan, and Xiaofei Ma.
\newblock Repoformer: Selective retrieval for repository-level code completion.
\newblock 2024.
\newblock URL \url{https://www.amazon.science/publications/repoformer-selective-retrieval-for-repository-level-code-completion}.

\bibitem[Zhang et~al.(2023)Zhang, Chen, Zhang, Keung, Liu, Zan, Mao, Lou, and Chen]{repocoder}
Fengji Zhang, Bei Chen, Yue Zhang, Jacky Keung, Jin Liu, Daoguang Zan, Yi~Mao, Jian-Guang Lou, and Weizhu Chen.
\newblock {R}epo{C}oder: Repository-level code completion through iterative retrieval and generation.
\newblock In Houda Bouamor, Juan Pino, and Kalika Bali, editors, \emph{Proceedings of the 2023 Conference on Empirical Methods in Natural Language Processing}, pages 2471--2484, Singapore, December 2023. Association for Computational Linguistics.
\newblock \doi{10.18653/v1/2023.emnlp-main.151}.
\newblock URL \url{https://aclanthology.org/2023.emnlp-main.151/}.

\bibitem[Zhang and Tan(2021)]{SplitterSparse2021}
Hang Zhang and Liling Tan.
\newblock Textual representations for crosslingual information retrieval.
\newblock In Shervin Malmasi, Surya Kallumadi, Nicola Ueffing, Oleg Rokhlenko, Eugene Agichtein, and Ido Guy, editors, \emph{Proceedings of the 4th Workshop on e-Commerce and NLP}, pages 116--122, Online, August 2021. Association for Computational Linguistics.
\newblock \doi{10.18653/v1/2021.ecnlp-1.14}.
\newblock URL \url{https://aclanthology.org/2021.ecnlp-1.14/}.

\bibitem[Zhang et~al.(2025)Zhang, Zhao, Wang, Yang, Wei, and Wu]{cAST2025}
Yilin Zhang, Xinran Zhao, Zora~Zhiruo Wang, Chenyang Yang, Jiayi Wei, and Tongshuang Wu.
\newblock cast: Enhancing code retrieval-augmented generation with structural chunking via abstract syntax tree, 2025.
\newblock URL \url{https://arxiv.org/abs/2506.15655}.

\bibitem[Zhou et~al.(2025)Zhou, Xie, Chen, He, and Li]{zhou2025}
Chunying Zhou, Xiaoyuan Xie, Gong Chen, Peng He, and Bing Li.
\newblock Multi-view adaptive contrastive learning for information retrieval based fault localization, 2025.
\newblock URL \url{https://arxiv.org/abs/2409.12519}.

\bibitem[Zhou et~al.(2023)Zhou, Alon, Xu, Jiang, and Neubig]{zhou2023docprompting}
Shuyan Zhou, Uri Alon, Frank~F. Xu, Zhengbao Jiang, and Graham Neubig.
\newblock Docprompting: Generating code by retrieving the docs.
\newblock In \emph{The Eleventh International Conference on Learning Representations}, 2023.
\newblock URL \url{https://openreview.net/forum?id=ZTCxT2t2Ru}.

\bibitem[Zhou et~al.(2024)Zhou, Kim, Xu, Liu, Han, and Lo]{LongTailImpact2024}
Xin Zhou, Kisub Kim, Bowen Xu, Jiakun Liu, DongGyun Han, and David Lo.
\newblock The devil is in the tails: How long-tailed code distributions impact large language models.
\newblock In \emph{Proceedings of the 38th IEEE/ACM International Conference on Automated Software Engineering}, ASE '23, page 40–52. IEEE Press, 2024.
\newblock ISBN 9798350329964.
\newblock \doi{10.1109/ASE56229.2023.00157}.
\newblock URL \url{https://doi.org/10.1109/ASE56229.2023.00157}.

\end{thebibliography}
\end{document}